\documentclass[journal]{IEEEtran}
\usepackage[utf8]{inputenc}

\usepackage[ruled, vlined]{algorithm2e}
\usepackage{multirow}
\usepackage{booktabs,dcolumn,array}           
\usepackage{verbatim}     
\usepackage{amsmath}
\usepackage{amssymb}
\usepackage{dsfont}
\usepackage{subcaption}
\usepackage{url}
\usepackage{hyperref}
\usepackage{lipsum}
\usepackage[normalem]{ulem}

\usepackage{xcolor,colortbl}
\usepackage{pgfplots}
\usetikzlibrary{pgfplots.groupplots}

\usepackage{cases}
\usepackage{graphicx}
\usepackage{booktabs}
\usepackage{multirow}
\usepackage{scrextend}

\pgfplotsset{compat=1.14}

\newlength\mylen
\newcommand\KwNewIn[1]{%
  \settowidth\mylen{\KwIn{}}%
  \setlength\hangindent{\mylen}%
  \hspace*{\mylen}#1\\}
\SetKwFor{RepTimes}{repeat}{times}{end}
\SetKwComment{Comment}{$\triangleright$\ }{}

\usepackage{etoolbox}
\makeatletter
\patchcmd{\@algocf@start}% <cmd>
  {-1.5em}{0pt}{}{}
\makeatother

\usepackage[skip=2pt,font=footnotesize]{caption}

\begin{document}
\title{Using Neural Networks for Fast SAR Roughness Estimation of High Resolution Images}

\author{Li Fan, and Jeova Farias Sales Rocha Neto %
\thanks{J. F. S. Rocha Neto is with the Department of Computer Science, Haverford College, Haverford, PA, 19041 USA e-mail: \url{jeovafarias@gmail.com}.}% <-this % stops a space
\thanks{Li Fan is with Department of Statistical Science, Duke University, Durham, NC, 27708, USA. e-mail: \url{li.fan@duke.edu}}.}%

\maketitle

\begin{abstract}
The analysis of Synthetic Aperture Radar (SAR) imagery is an important step in remote sensing applications, and it is a challenging problem due to its inherent speckle noise. One typical solution is to model the data using the $G_I^0$ distribution and extract its roughness information, which in turn can be used in posterior imaging tasks, such as segmentation, classification and interpretation. This leads to the need of quick and reliable estimation of the roughness parameter from SAR data, especially with high resolution images. Unfortunately, traditional parameter estimation procedures are slow and prone to estimation failures. In this work, we proposed a neural network-based estimation framework that first learns how to predict underlying parameters of $G_I^0$ samples and then can be used to estimate the roughness of unseen data. We show that this approach leads to an estimator that is quicker,  yields less estimation error and is less prone to failures than the traditional estimation procedures for this problem, even when we use a simple network. More importantly, we show that this same methodology can be generalized to handle image inputs and, even if trained on purely synthetic data for a few seconds, is able to perform real time pixel-wise roughness estimation for high resolution real SAR imagery. 
\end{abstract}

% Note that keywords are not normally used for peerreview papers.
\begin{IEEEkeywords}
Synthetic Aperture Radar Images, Neural Networks, Image Analysis, Statistical Modeling.
\end{IEEEkeywords}

\section{Introduction}

\IEEEPARstart{S}{ynthetic} Aperture Radar (SAR) data is crucial to remote sensing and earth monitoring. Its ability to capture high resolution snapshots of targets and landscapes independently of the weather conditions and sunlight has opened the way to important advancements in environmental monitoring, emergency response,  evaluation of damages in natural catastrophes, urban planning and ecology, to mention a few. Because of its widespread use, there is a considerable industrial appeal for SAR image understanding algorithms. This task, however, is challenging due to the degrading speckle noise inherent to such data, which prevents the application of image processing techniques that are common in other image domains. 

Thankfully, this same appeal led to the developments of statistical models that describe the SAR data despite its noise pattern. In particular, the $G_I^0$ \cite{frery1997model} distribution for intensity data, highly successful in both theory and practice \cite{frery2022sar}, arouse to prominence in the remote sensing community because of its capacity to model well textured, extremely textured and textureless terrain data. To do so, this model relies on three parameters, one of which is roughness. It directly corresponds to the captured texture and has been extensively used for SAR data understanding \cite{neto2019level, cassetti2022improved, gambini2015parameter, khan2013application, rodrigues2016sar}. Estimating such parameter became crucial in many SAR imaging techniques \cite{rodrigues2016sar, cassetti2022improved}. However, the best current available algorithms for such a task, i.e. Maximum Likelihood \cite{frery2022sar}, Log-cumulant \cite{nicolas2002introduction} and Minimum Distance Estimators \cite{gambini2015parameter}, are too computationally expensive to be applied to high resolution images  within reasonable time.

Taking a different approach, deep learning has been successfully applied to SAR \cite{zhu2021deep} and many other imaging domains \cite{minaee2021image}. In these techniques, one typically relies on training a large neural network on a substantial amount of labeled data, or augmentations thereof. Practically speaking, this considerably sized process can be  accomplished by implementing parallelizable learning techniques in powerful parallel processing machines, such as Graphics Processing Units (GPUs). Nowadays, many efficient and simple-to-use libraries are available for manipulating data, and training deep networks on GPUs, which prompts researchers and practicians alike to design their methodologies in a way to leverage these powerful tools.

In this work, we propose using neural network-based learning algorithm for estimating  roughness maps in SAR images. In our methodology, we generate sets of $G_I^0$ samples of varying sizes, compute their sample moments and use them to predict the parameters that generated the data. Once the network is trained on this fully synthetic data, we proceed with network inference on real data.
% , which in this case leads to parameter estimation.  
To the best of our knowledge, our work is the first of its kind to train a neural network to estimate parameters of distributions based on sample moments from synthetic data. As we shall show, in fact, a very small network composed, trained for a few seconds on a small dataset, is sufficient to produce better estimates on unseen data than maximum likelihood and log-cumulant-based estimators in terms of mean squared error, while being also faster and less prone to failures. Finally, this same process can be fully implemented, from sample moment computation and training to network inference, on a GPU for high resolution image-sized input, with pixel-wise roughness estimation accomplished in a few milliseconds on average. Overall, we propose a simple, principled and efficient method for training neural networks for SAR image understanding, using the statistical baggage this type of data carries.

% This paper unfolds as follows. In Section II, we review SAR image modeling and Neural Networks. In Section III, we lay out our proposed estimation methodology, initially treating the data as a collection of samples organized in no special way, and then assuming they are in an image grid. We then provide some numerical experimental results to demonstrate the performance of our algorithms in Section IV. In Section V, we summarize our findings and conclude our work.

\section{Preliminaries}

\subsection{SAR Statistical Modeling and Parameter Estimation}\label{sec:sarintro}

% \textcolor{orange}{\sout{Because of the nature of a radar system, t}T}
The intensity return $Z$ in monopolarized SAR image can be effectively modeled by the product of two independent random variables: $X$, the \textit{backscatter data}, and $Y$, the inherent \textit{speckle noise} \cite{frery2022sar}. Typically, $Y$ is modeled as a unitary-mean Gamma distributed random variable with shape parameter corresponding to the number of looks $L \in \mathbb{N}^*$ used to capture the data, and considered known during estimation \cite{frery2022sar}. Assuming that $X$ obeys the reciprocal of Gamma law, Frery\textit{ et Al}. \cite{frery1997model} showed that $Z \sim G_I^0(\alpha, \gamma, L)$ with density:
\begin{equation}\label{gi0}
  f_{\mathcal{G}^{0}_I}(z,\theta ) = \frac{L^{L}\Gamma(L-\alpha)}{\gamma^{\alpha}\Gamma(-\alpha)\Gamma(L)} z^{L-1}(\gamma+Lz)^{\alpha - L},
\end{equation}
where $-\alpha, \gamma > 0$ are called \textit{roughness} and scale, respectively.

The $G_I^0$ model has become a popular choice for SAR data due to its expressiveness \cite{frery1997model}, mathematical tractability \cite{frery2022sar}, and effectiveness when applied to various imaging tasks \cite{nobre2016sar, cassetti2022improved}. In fact, it was shown that the roughness parameter $\alpha$ plays an important role in SAR understanding \cite{gambini2015parameter}. With roughness close to zero, typically $\alpha > -3$, it suggests the targeted region is highly textured, therefore urban. For $\alpha \in [-3, -6]$, we have evidence for moderately textured regions, which can correspond to forests. Finally,  $\alpha  < -6$ implies textureless areas, such as seas and pastures. Hence, using $\alpha$ estimates instead of pixel intensities on SAR images led to improvements in image segmentation \cite{rodrigues2016sar, neto2019level} and region discrimination \cite{nascimento2009hypothesis}.

This direct application of SAR statistical modelling fueled the study and improvement of estimators for the $G_I^0$ distribution \cite{nascimento2009hypothesis, gambini2015parameter}. While early work employed a Maximum Likelihood Estimator (MLE) \cite{frery2022sar} for parameter estimation, later developments led to the application of Method of Moments \cite{frery2022sar}
%\cite{cheng2013improved}
, Method of Log Cumulants (LCUM) \cite{nicolas2002introduction, khan2013application, rodrigues2016sar} and Minimum-Distance Estimators  \cite{gambini2015parameter, cassetti2022improved} to this problem. Despite some of their successes, all these methods, especially LCUM, heavily rely on slow optimization procedures and are prone to high estimation failures \cite{gambini2015parameter}. These issues hinder their application in image-sized data understanding, as the current praxis relies on generating \textit{roughness maps} by sweeping the image with a window that collects the intensity data centered at each pixel and then proceeds with the parameter estimation algorithm \cite{rodrigues2016sar}, a process that can be excessively slow.

\subsection{Neural Networks}\label{sec:neural_net_prelim}
We focus on supervised learning algorithms that, given a dataset $\mathcal{D} = \{(\mathbf{x}_i, \mathbf{y}_i)\}_{i = 1}^N$, aim to compute a model $f_\theta(\cdot)$ parameterized by $\theta$ that best predicts the output $\mathbf{y}_i$'s when given the input $\mathbf{x}_i$'s. The prediction error, to be minimized, is measured via a loss, typically the Mean Squared Error (MSE):
% The distance between the prediction and the truth, to be minimized, is measured by the value of a loss function, typically the Mean Squared Error (MSE):
\begin{equation}\label{eq:mse}
    \operatorname{MSE} = \frac{1}{N}\sum_{(\mathbf{x}, \mathbf{y}) \in \mathcal{D}}\lVert \mathbf{y} - f_{\theta}(\mathbf{x})) \rVert^2_2,
\end{equation}
where $\lVert \cdot \rVert_2$ is the Euclidean vector norm.  In \textit{Neural Networks}, we assume that $f_{\theta}$ is composed of units of a non-linear activations $\sigma(\cdot)$ and parameters $\theta = \{\mathcal{W}, \mathcal{B}\}$, where $\mathcal{W} = \{W_i\}_{i=1}^Q$ are \textit{weight} matrices  and $\mathcal{B} = \{\mathbf{b}_i\}_{i=1}^Q$ a set of \textit{bias} vectors. The Multilayer Perceptron (MLP), a popular network, uses a model $NN_{\theta}(\mathbf{x})$ formed by the composition of affine transformations of the inputs followed by the element-wise application of $\sigma(\cdot)$: 
\begin{equation}
    NN_{\theta}(\mathbf{x}) = \sigma(W_{Q}\ldots \sigma(W_{2}\sigma(W_1\mathbf{x} + \mathbf{b}_1)+\mathbf{b}_2)\ldots + \mathbf{b}_{Q}).
\end{equation}
Each transformation is typically referred to as a \textit{layer} of the MLP, where the first and last are the input and output layers, and  others are called hidden layers. A network with multiple hidden layers is commonly known as a ``deep'' neural network.

Another, arguably more popular, neural network model is the Convolutional Neural Network (CNN). Different from the MLP model, in CNN, the weights in each layer are shared so that they effectively perform a convolution operation  on the input. For example, consider an image $I$ of shape $c \times w \times l$, where $w$ and $l$ are 2D spatial dimensions of $I$, and  $c$ is its number of channels (three for RGB data). One can design a matrix of weights whose application on a vectorized/flattened version of $I$ corresponds to a convolution of $I$ on its original shape and a \textit{kernel matrix} of shape $c \times k \times k$, where $k \ll w, l$. Using such \textit{convolutional layers} brings several advantages for learning on image-sized data. First, the number of weights to be learned is drastically reduced, from $O(cwl)$ to $O(ck^2)$. Second, the number of weights to be learned no longer depends on the size of the input. Third, one can show that this process learns separate feature detectors, which improves the network performance on important imaging tasks such as detection, denoising and segmentation \cite{minaee2021image}.

More importantly to us is the setting where MLPs and CNNs are equivalent. Suppose we have the input vectors of the dataset $\mathcal{D}$ in $\mathbb{R}^d$ and output vectors in $\mathbb{R}^{d'}$. Learning a dense layer for  $\mathcal{D}$ would require finding a matrix $W \in \mathbb{R}^{d \times d'}$. Now, consider the tensors $X \in \mathbb{R}^{d \times w \times l}$ and $Y \in \mathbb{R}^{d' \times w \times l}$, where $w$ and $l$ are chosen such that $wl = N$, and each $\mathbf{x}_n$ (resp. $\mathbf{y}_n$) is placed in a transversal tube $X[:, i, j]$ (resp. $Y[:, i, j]$)\footnote{We use a notation similar to Numpy's for array indexing (for example, $X[i, j]$ is the element in $X$ at indices $(i, j)$) and slicing (or example, $X[i, :]$ is all the elements in the $i$-th row of $X$).}\cite{kolda2009tensor}. It is easy to see that training $d'$ kernel matrices of shape $d \times 1 \times 1$  is the same as learning the values in $W$ \cite{lin2013network}. As discussed in the next section, using these $1 \times 1$ convolutional layers will enable us to see our parameter estimation procedure as an inference on a pre-trained CNN, leveraging the speed of highly parallel computing engines during estimation and preventing us from using time-consuming loops for generating roughness maps.

\section{Proposed Methodology}
\subsection{Estimating $G_I^0$ Using Neural Networks}
Our approach for estimation relies on neural network-based learning procedure. Let $L \in \mathbb{N}^*$ be known \cite{frery2022sar}
%\footnote{We follow the traditional literature and  assume $L$ is known beforehand  , although its estimation could be included in our framework.} 
and $\mathcal{H} = \{(\mathcal{Z}_i, \alpha_i)\}_{i = 1}^N$ be a dataset composed of samples $\mathcal{Z}_i$ from a $G_I^0 (\alpha_i, \gamma_i, L)$ distribution of varying sizes, where we assume $\gamma_i = - \alpha_i -1$ \footnote{We follow the praxis of assuming that the samples have unit mean \cite{frery2022sar}.}.  We also assume that $\alpha_i$ is  always above a given lower bound (set to $-15$ in our experiments) \cite{gambini2015parameter}. 

Our goal is to train a neural network to predict $\alpha_i$ from $\mathcal{Z}_i$.  However, the data in $\mathcal{Z}_i$ cannot be used directly, as the input vector of our network needs to be of fixed size. The final estimator also needs to be invariant to sample permutations in each input set. To solve these issues, the authors in \cite{zaheer2017deep} proved that that all permutation invariant functions $t(\cdot)$ on a set $\mathcal{S}$ can be represented as $t(\mathcal{S}) = \rho\left(\sum_{s \in \mathcal{S}} \phi(s)\right)$, where $\rho(\cdot)$ and $\phi(\cdot)$ are continuous functions. Setting $\rho(x) = \frac{1}{|\mathcal{Z}_i|}x$ and $\phi(x) = (\log x)^m$ and applying the resulting $t(\cdot)$ on a sample $\mathcal{Z}_i$ from $\mathcal{D}$, we get:

\begin{equation}\label{eq:sampleLCum}
\mu_m(\mathcal{Z}_i) = \frac{1}{|\mathcal{Z}_i|}\sum_{z \in \mathcal{Z}_i} (\log z)^m,
\end{equation}
where $\mu_m(\mathcal{Z}_i)$ conveniently represents the sample log-moment of order $m$, a statistic commonly used in SAR literature \cite{khan2013application, rodrigues2016sar}. In practice, we create a vector of $N_m$ moments $\boldsymbol{\mu}_i = [\mu_1(\mathcal{Z}_i)\,\, \mu_2(\mathcal{Z}_i)\,\, \ldots \,\, \mu_{N_m}(\mathcal{Z}_i)]$ and train our network on the dataset $\mathcal{D} = \{(\boldsymbol{\mu}_i, \alpha_i)\}_{i = 0}^N$.  We use an MLP as our network architecture, since its structure can be adapted to infer full sized roughness maps, as explained in the next section.  

\begin{algorithm} \small
\caption{\small Training Roughness Estimator}\label{alg:rougness_samples}
\KwIn{$L$: \# looks. $R$: dataset size. $T$: \# epochs. $N_m$:  \# log-mom.}
\KwNewIn{$\mathcal{A}$: set of $\alpha$ values, $\mathcal{S}$: set of sample sizes.}
% \KwNewIn{$\mathcal{S}$: set of sample sizes to be trained on.}

% \KwOut{$NN_{\theta}$: A roughness estimator for a $G_I^0$ sample.}

  \SetKwFunction{FMain}{Main}
  \SetKwFunction{FTrainDS}{SyntheticDataset}
  \SetKwFunction{FMom}{ComputeMoments}
  \SetKwFunction{FTrain}{TrainNeuralNet}

    \SetKwProg{Fn}{Function}{:}{\KwRet $NN_{\theta}$}
    \Fn{\FMain{}}{
        $\mathcal{D} \gets $ \FTrainDS{$L, w, h, \mathcal{A}, \mathcal{K}, N_m, R$}
        
        $NN_{\theta} \gets$ \FTrain{$\mathcal{D}, T$}
    }

    \SetKwProg{Fn}{Function}{:}{\KwRet $\mathcal{D}$}
    \Fn{$\mathcal{D}=$ \FTrainDS{$L, \mathcal{A}, \mathcal{S}, N_m, R$}}{
        $\mathcal{D} \gets $ Empty set 
        
        \RepTimes{R}{
            $\alpha, s \gets $ Sample uniformly from $\mathcal{A}$ and $\mathcal{S}$, resp.
            
            $\gamma \gets -\alpha -1$

            $\mathcal{Z} \gets \{z | z \sim G_I^0(\alpha, \gamma, L), |\mathcal{Z}| = s\}$, 
      
            $\boldsymbol{\mu} \gets $ \FMom{$\mathcal{Z}, N_m$}
            
            $\mathcal{D} \gets \mathcal{D} \cup \{(\boldsymbol{\mu}, \alpha)\}$
        }
            }

        \SetKwProg{Pn}{Function}{:}{\KwRet $X$}
    \Pn{$\boldsymbol{\mu}= $ \FMom{$\mathcal{Z}, N_m$}}{
        % $\boldsymbol{\mu} \in \mathbb{R}^{N_m}$
        
        \ForEach{$m \in \{1, 2, \ldots, N_m\}$}{
            $\boldsymbol{\mu}[m] \gets \frac{1}{|\mathcal{Z}|}\sum_{z \in \mathcal{Z}} (\log z)^m$\,
        }
    }

    \SetKwProg{Fn}{Function}{:}{\KwRet $NN_{\theta}$}
    \Fn{$NN_{\theta}=$ \FTrain{$\mathcal{D}$, $T$}}{
        $NN_{\theta} \gets$ Multilayer Perceptron with parameters $\theta$.
        
        \RepTimes{T}{
            \ForEach{\textnormal{batch} $(\mathbf{x}, \mathbf{y}) \in \mathcal{D}$}{
                Optimize $\theta$ to minimize $\lVert \mathbf{y} - NN_{\theta}(\mathbf{x}) \rVert_2^2$ 
            }
        }
    }
    
\end{algorithm}

\begin{algorithm} \small
\caption{\small Training Roughness Map Estimator}\label{alg:rougness_image}
\KwIn{$L$: \# looks. $w$ and $h$: image dimensions.}
\KwNewIn{$R$: dataset size, $T$: \# epochs. $N_m$: \# log-moments.}
% \KwNewIn{$w$ and $h$: synthetic image dimensions,}
\KwNewIn{$\mathcal{A}$: set of $\alpha$ values, $\mathcal{K}$:  set of kernel sizes,}
% \KwNewIn{$\mathcal{K}$: set of kernel sizes}
% \KwNewIn{$T$: number of training epochs.}
% \KwOut{$NN_{\theta}$: A roughness map estimator for an image.}
    \SetKwFunction{FMain}{Main}
    \SetKwFunction{FTrainDS}{SyntheticDataset}
    \SetKwFunction{FMom}{ComputeImgMoments}
    \SetKwFunction{FTrain}{TrainNeuralNet}

    \SetKwProg{Fn}{Function}{:}{\KwRet $NN_{\theta}$}
    \Fn{\FMain{}}{
        $\mathcal{D} \gets $ \FTrainDS{$L, w, h, \mathcal{A}, \mathcal{K}, N_m, R$}
        
        $NN_{\theta} \gets$ \FTrain{$\mathcal{D}, T$}
    }
    
    \SetKwProg{Fn}{Function}{:}{\KwRet $\mathcal{D}$}
    \Fn{$\mathcal{D} =$ \FTrainDS{$L, w, h, \mathcal{A}, \mathcal{K}, N_m, R$}}{
        $\mathcal{D} \gets $ Empty set 
        
        \RepTimes{R}{
                $\alpha, k \gets $ Sample uniformly from $\mathcal{A}$ and $\mathcal{K}$, resp.     
                
                $\gamma \gets -\alpha -1$

                $I \gets \text{Matrix in } \mathbb{R}^{w \times h}, \text{with } I[i, j] \sim G_I^0(\alpha, \gamma, L)$
                
                $A \gets  \text{Matrix in } \mathbb{R}^{w \times h}, \text{with } A[i, j] = \alpha$

                $M \gets $ \FMom{$I, N_m, k, \alpha, \gamma, L$}

                $\mathcal{D} \gets \mathcal{D} \cup \{(M, A)\}$
            
        }
    }

    \SetKwProg{Pn}{Function}{:}{\KwRet $M$}
    \Pn{$M =$ \FMom{$I, N_m, k, \alpha, \gamma, L$}}{

        $I \gets I + $padding of $G_I^0(\alpha, \gamma, L)$ samples.
        
        % $M \gets \text{Tensor in } \mathbb{R}^{N_m \times w \times h}$
        
        \ForEach{$m \in \{1, 2, \ldots, N_m\}$}{
            $M[i, :, :] \gets \operatorname{AvgPool}(\log(I)^m, k \times k \text{ kernel})$\,
        }
    }

    \SetKwProg{Fn}{Function}{:}{\KwRet $NN_{\theta}$}     
    \Fn{$NN_{\theta}=$ \FTrain{$\mathcal{D}$, $T$}}{
        $NN_{\theta} \gets$ FCNN with only $1 \times 1$ convolutions.
        
        \RepTimes{T}{
            \ForEach{\textnormal{batch} $(X, Y) \in \mathcal{D}$}{
                Optimize $\theta$ to minimize $\lVert Y - NN_{\theta}(X) \rVert_2^2$ 
            }
        }
            }
\end{algorithm}

Algorithm \ref{alg:rougness_samples} gives an overview of this training procedure. Following  \texttt{Main()}, a synthetic $G_I^0$ dataset is created by drawing samples of various predefined sample sizes and $\alpha$ values. For each set, \texttt{ComputeMoments()} runs on each sample to find its first $N_m$ log-moments and concatenate the resulting moment vector to the roughness value that generated it. Finally, an MLP is trained on these pairs of moment vectors and the $\alpha$ parameter over a given amount of epochs.

After the network is trained, one can estimate the parameters of a given unseen sample set by (1) computing its moments via  \texttt{ComputeMoments()} and (2) feeding them through the trained model. We hope that, although training requires extra computational time, the inference step in this process is fast.

\subsection{Adaptation to Roughness Map Estimation}
In many practical scenarios involving parameter estimation in SAR, one wishes to estimate the roughness of all pixel locations in a potentially high resolution image $I$ \cite{rodrigues2016sar}. While we could apply the algorithm described  previously to all $k\times k$ sized windows in $I$ for a given $k>0$, in this section we show the functions described in Algorithm \ref{alg:rougness_samples} can be easily adapted to image size data and be fully implemented on a GPU. This transition will enable us to fully parallelize our estimation algorithm in both training and inference phases, which will consequently highly reduce our computation time.

In this new setting, let $I \in \mathbb{R}^{w \times h}$. For our  training phase, each input in our training set will be composed of samples from the $G_I^0$ distribution for unique values of $\alpha$, $\gamma$ and $L$ disposed in a $w \times h$ matrix. In the inference phase, $I$ is a real SAR image of any size, meaning that our method will be able to estimate the roughness of an image with a network trained on purely synthetic and random data.

Our goals are (1) to efficiently compute all the desired $N_m$ moments of the data surrounding each pixel in $I$ on a $k \times k$ window, composing a tensor $M \in \mathbb{R}^{N_n \times w\times h}$, and (2) to feed this data to an appropriate network that estimates the roughness parameters of each pixel location on the image grid. For step (1), one can use an Average Pooling Layer, $\operatorname{AvgPool}(\cdot)$, where a convolution kernel sweeps the input data, averaging the pixel values within that window \cite{bieder2021comparison}. Here we add an appropriate padding composed of $G_I^0$ samples to $\operatorname{AvgPool}$'s input, so its output size remains the same. Now, if the input data is $\log(I)^m$, where the exponentiation is computed pixel-wise, one can estimate each pixel's log-moments of order $m$ via $\operatorname{AvgPool}(\log(I)^m)$. Each channel in $M$ is computed by applying this procedure to all $m \in \{1, 2, \cdots, N_m\}$.

For step (2), we can use the connection between MLPs and $1\times 1$ convolutional networks described in Section \ref{sec:neural_net_prelim} and turn the MLP architecture used in Algorithm \ref{alg:rougness_samples} in to a Fully Convolutional Neural Network (FCNN). In that network, the input of its convolutions will be the transversal tubes on $M$, which correspond to the sample log-moments of each pixel in $I$. During training, its output will be a matrix $A \in \mathbb{R}^{w \times h}$, such that $A[i, j] = \alpha, \forall i,j$, where $\alpha$ is the roughness parameter that generated $I$ along with $\gamma = -\alpha -1$ and a predetermined $L$. Algorithm \ref{alg:rougness_image} explains this training algorithm in more detail and is analogous to Algorithm \ref{alg:rougness_samples} in its execution.

\section{Numerical Experiments and Discussion}

\begin{figure}[t]
\centering
 \includegraphics[width= \linewidth]{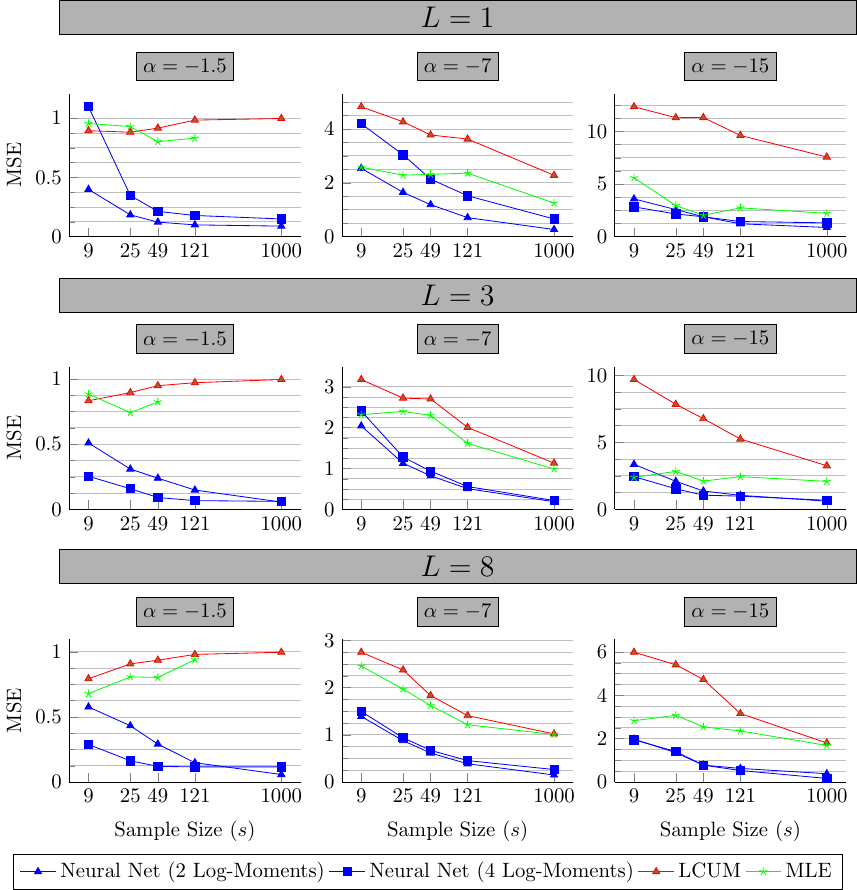}
 \caption{MSE values for the synthetic data. Missing values on graphs correspond to the method failing in all simulations.}\label{fig:MSE_Gi0}
\end{figure}

\subsection{Data and Algorithmic Setup and Assessment Methodology}
We assess the performance of our proposed methodology on two settings. The first is a synthetic one where we aim at estimating the roughness parameter from samples of the $G_I^0$ distribution. We qualitatively compare the resulting network from Algorithm \ref{alg:rougness_samples} to the standard estimators for this problem. In our second experiment, we qualitatively evaluate a model trained using Algorithm \ref{alg:rougness_image} on a real SAR image.

To sample synthetic SAR data from the $G_I^0$ model, we make use of its multiplicative nature and sample $X \sim \Gamma(1, L)$ and $Y' \sim \Gamma(-\alpha, \gamma)$ to get $Z = X/Y' \sim G_I^0(\alpha, \gamma, L)$ \cite{frery2022sar}. In our real experiments, we use a $1500 \times 1500$ SAR image acquired by Airborne SAR (AIRSAR) in HH polarization with $L = 1$ in C-band over Lake Superior, near Sault Saint Marie, MI.

We choose simple and analogous network architectures to be trained in both Algorithms \ref{alg:rougness_samples} and \ref{alg:rougness_image}. The MLP in Algorithm \ref{alg:rougness_samples} consists of two hidden layers, one with eight and the other with four units, followed by $\tanh()$ activation functions. The FCNN in Algorithm \ref{alg:rougness_image} is composed of two $1\times 1$ convolutional layers with eight and four filters, respectively, followed again by $\tanh()$ activations. 
These activations were chosen as they performed better than the traditional ReLU activation. For experimental simplicity, no batch normalization, dropout or data preprocessing were used during training. 

\begin{figure*}
    \begin{subfigure}{.18\linewidth}
        \captionsetup{justification=centering} 
        \includegraphics[width=\textwidth]{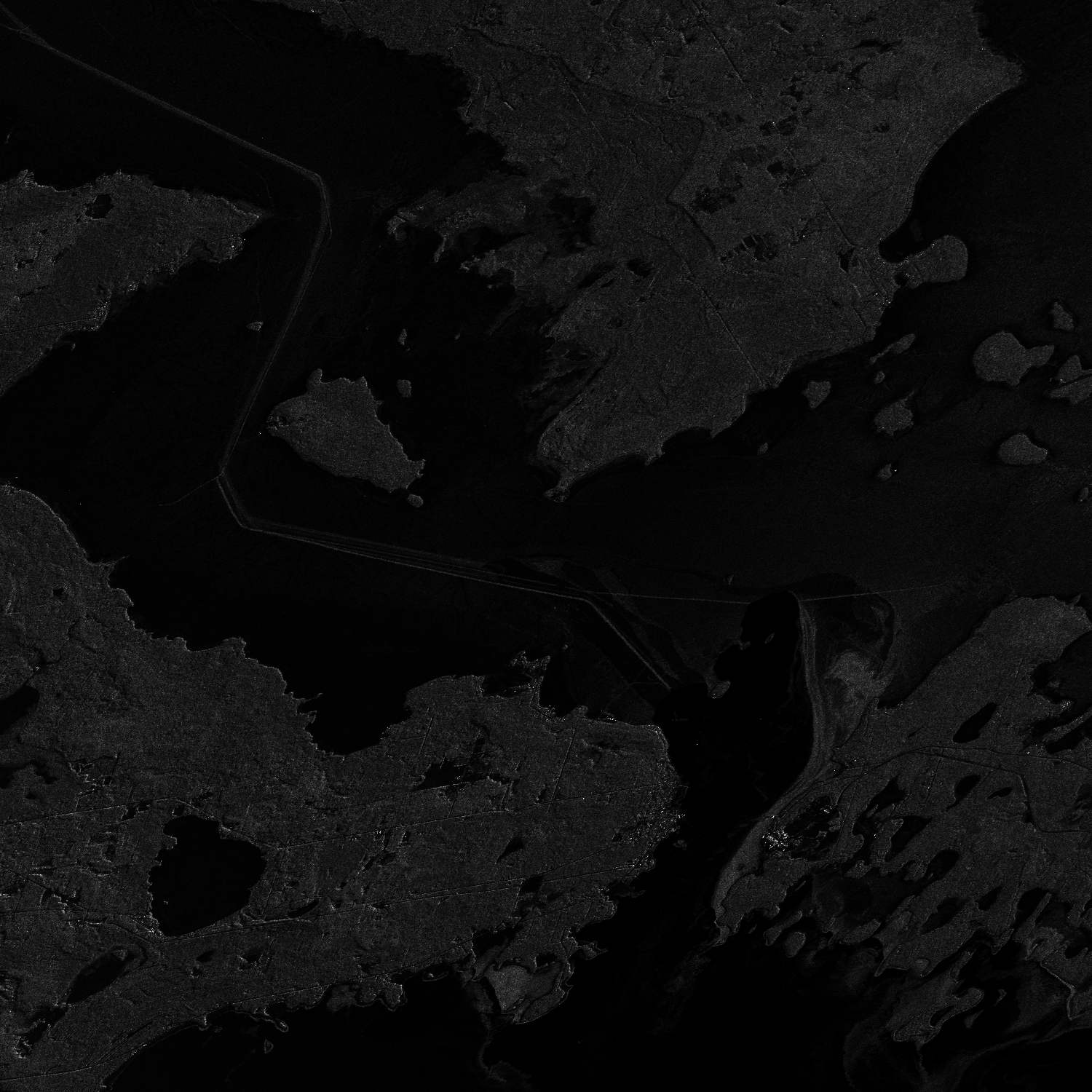}
        \caption{Original Image}
    \end{subfigure}%
    \hfill
    \begin{subfigure}{.18\linewidth}
        \captionsetup{justification=centering} 
        \includegraphics[width=\textwidth]{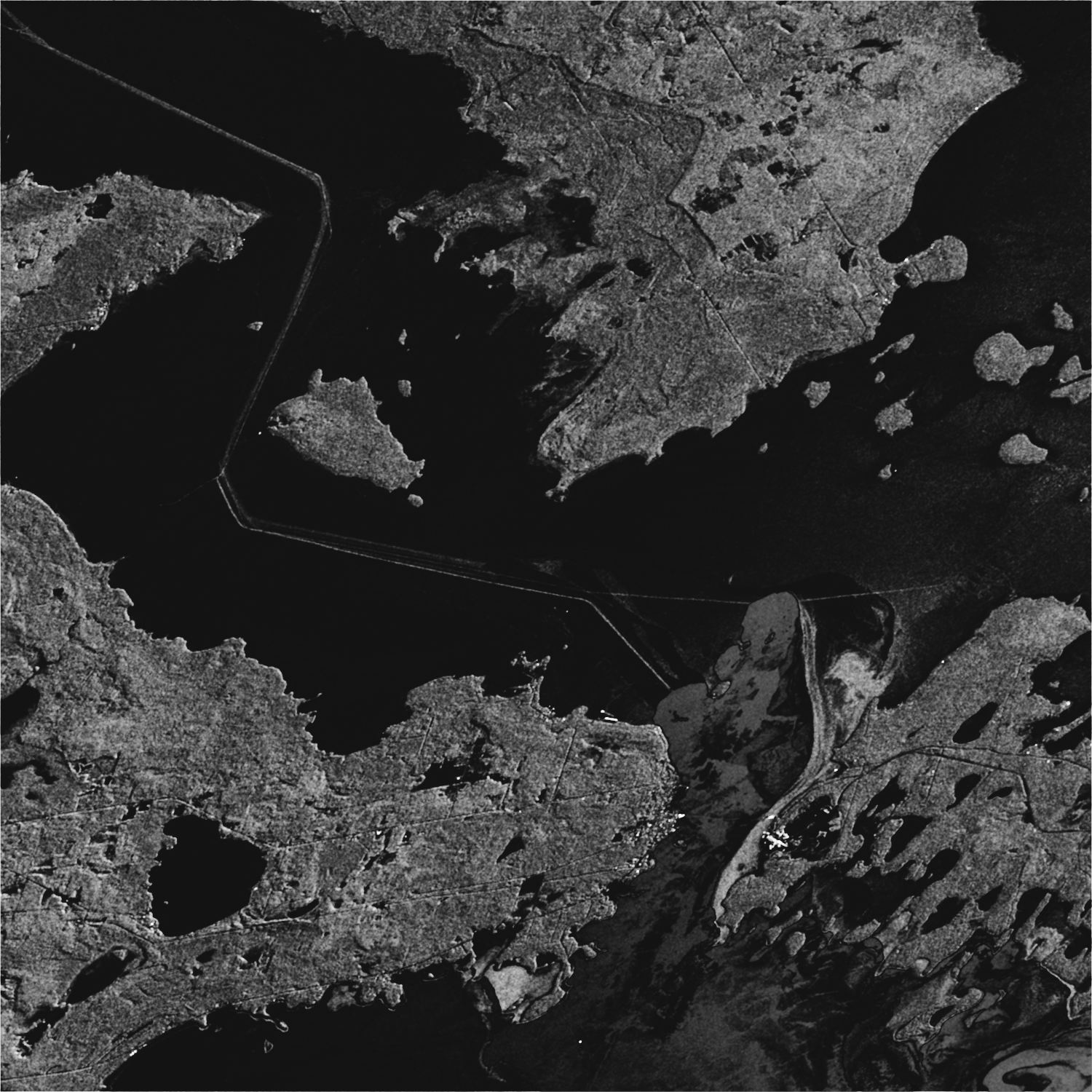}
        \caption{$k = 3$ (0.097 s)}
    \end{subfigure}%
    \hfill
    \begin{subfigure}{.18\linewidth}
        \captionsetup{justification=centering} 
        \includegraphics[width=\textwidth]{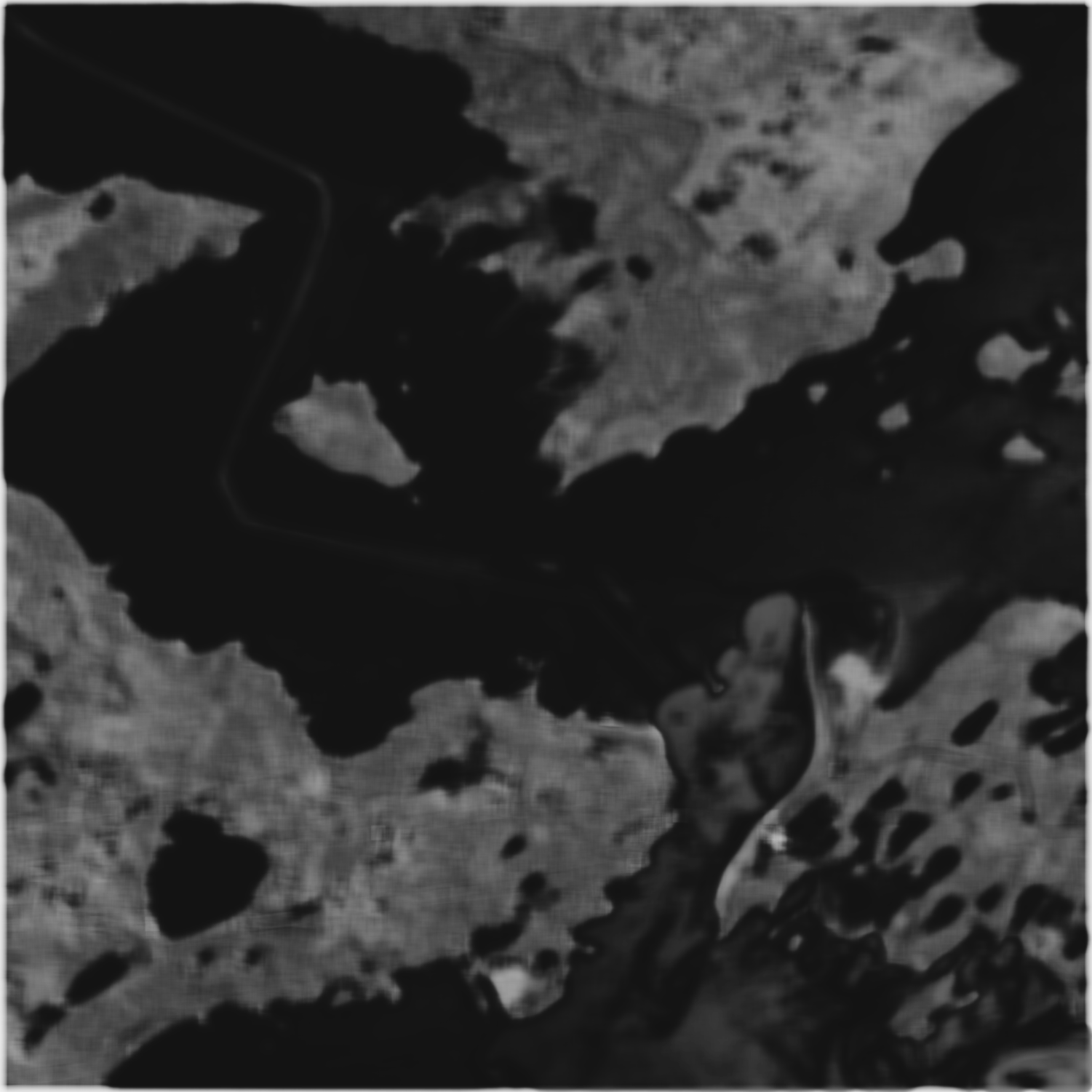}
        \caption{$k = 11$  (0.637 s)}
    \end{subfigure}%
    \hfill
    \begin{subfigure}{.18\linewidth}
        \captionsetup{justification=centering} 
        \includegraphics[width=\textwidth]{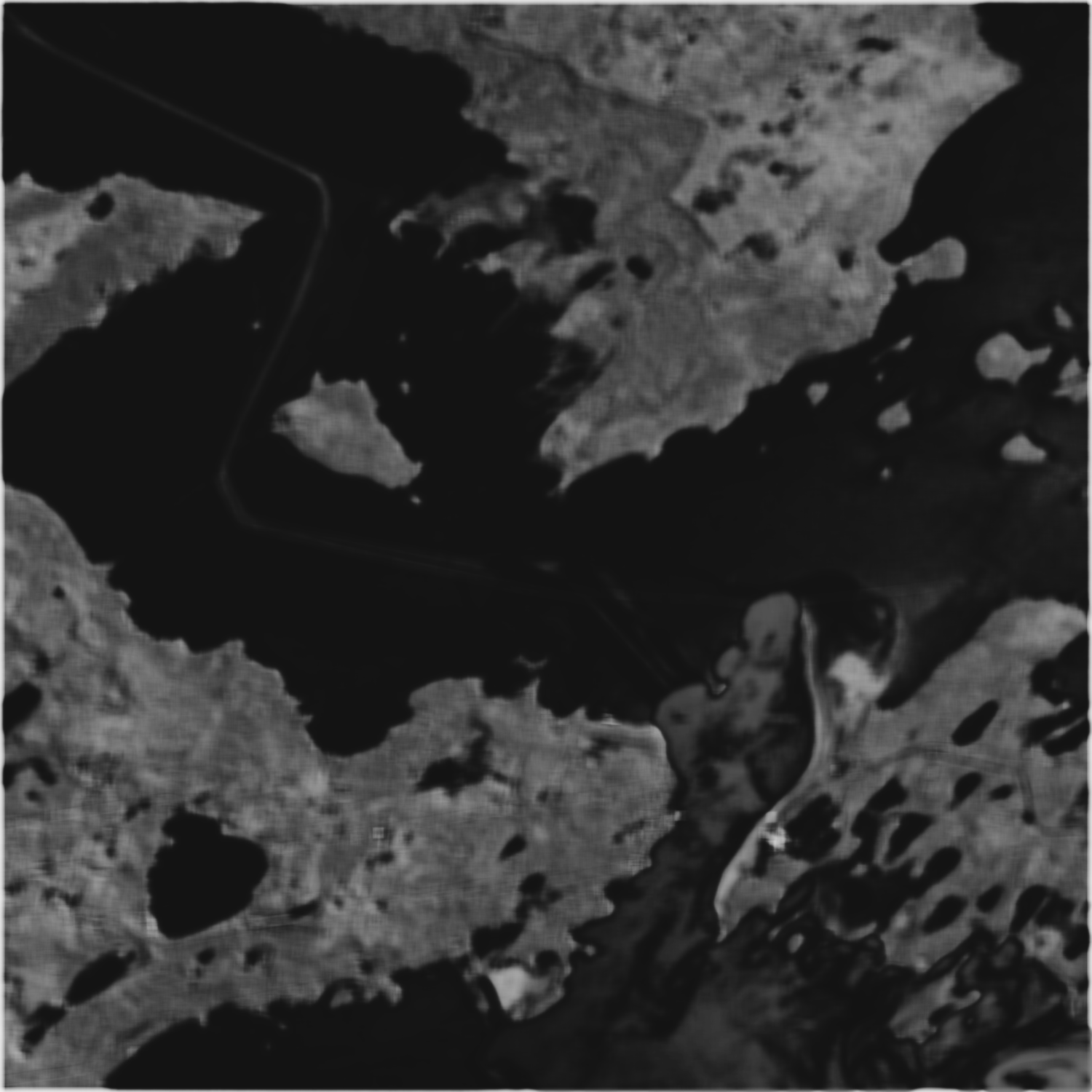}
        \caption{$k = 20$  (2.206 s)}
    \end{subfigure}%
    \hfill
    \begin{subfigure}{.224\linewidth}
        \captionsetup{justification=centering} 
        \includegraphics[width=\textwidth]{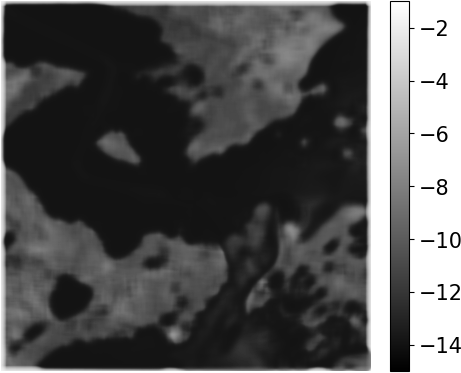}
        \caption{$k = 45$ (11.887 s)}
    \end{subfigure}%
    \caption{Qualitative results on a real SAR image. (a) The original intensity image. (b)-(e) The estimated roughness maps for different kernel sizes along with their respective elapsed computation time. The last color bar is shared by all maps. The images are better visualized in the digital version of this paper.} \label{fig:real_results}
\end{figure*}

\begin{table}[b]
\caption{Failure rates (\%) on synthetic data.}\label{fig:table_fail}
\centering
\begin{tabular}{ccccc}  
\toprule[1pt]
\multirow{2}{*}{$L$} & \multirow{2}{*}{MLE} & \multirow{2}{*}{LCUM} & \multicolumn{2}{c}{Neural Network}\\
& & & 2 Log-moments & 4 Log-moments\\\midrule
$1$    & 71.84     & 37.44      & 0.14     & 3.24 \\
$3$    & 59.84      & 25.55      & 1.46     & 6.22 \\
$8$    & 47.99       & 18.59      & 6.04     & 6.96 \\
\bottomrule[1pt]
\end{tabular}
\end{table}

Our networks were trained following the predicament in the \texttt{TrainNeuralNet()} functions in both Algorithms \ref{alg:rougness_samples} and \ref{alg:rougness_image} using Adaptive Moment Estimation (ADAM) optimizer \cite{kingma2014adam} with learning rate of 0.001 and batch size of 32. For Algorithm \ref{alg:rougness_samples}, we set $\mathcal{S} = \{100, 1000, 10000\}$ and tested $N_m = 2$ and $4$, and for Algorithm \ref{alg:rougness_image}, we have $(w, h) = (10, 10)$ and $\mathcal{K} = \{2, 5, \ldots, 11\}$ and only used $N_m = 2$. Both algorithms use $\mathcal{A} = \{-15, -13.5, \cdots, -1.5\}$, $R = 1000$ datapoints and $T = 300$ epochs, enough for their training convergence.

We compare our algorithms to LCUM and MLE results. To optimize both methods, we use Scipy's 
%\cite{virtanen2020scipy} 
\texttt{fsolve}, a Python wrapper for an implementation of Powell's dog leg method \cite{more1980user}. We start each optimization with $\alpha = -1.0001$. 

In our synthetic experiments, the quantitative assessment of our methods' estimation performance compared the estimated roughness with their ground truth values via Mean Square Error (MSE), as depicted in \ref{eq:mse}. For each experiment, we also compared the failure rates for each algorithm, where we consider an $\alpha$ estimate to fail if (1) it is not in the interval $[-1.5, -15]$ \cite{gambini2015parameter}, or (2) if its optimization procedure, in the case of MLE and LCUM, did not converge. 

We used a Tesla T4(R) GPU with 16Gb of RAM for our neural network training. The other methods were run on an Intel(R) Xeon(R) CPU at 2.30GHz with 26Gb of RAM. All methods were implemented in Python 3 and Pytorch \footnote{The code used to generate the results from this paper can be found at \url{https://github.com/jeovafarias/SAR-Roughness-Estimation-Neural-Nets.git}.}. Training each network took around 30 seconds each.

\subsection{Results on Synthetic Data}
We start by evaluating our estimation methods on synthetic samples of size $s \in \{9, 25, 49, 121, 1000\}$,  generated using roughness $\alpha \in \{-1.5, -7, -15\}$ and number of looks $L \in \{1, 3, 8\}$. For each setting, we performed a Monte Carlo experiment with 1000 $G_I^0$ samples. Figure \ref{fig:MSE_Gi0} compares MLE, LCUM and the networks trained using Algorithm \ref{alg:rougness_samples} with 2 and 4 log-moments when estimating $\alpha$ for each sample. We only considered the samples whose estimation did not fail to compute the MSE values. Note that we run our networks on sample sizes that they were not trained on and, despite that, both of them outperform its counterparts in most scenarios. The one trained on fewer log-moments generally performed better, potentially corroborating the premise in LCUM that also only uses two log-moments in its algorithm. This also means that our neural network-based approach more competently utilizes the information contained in those log-moments for estimation than LCUM, especially for lower $\alpha$ values.

Table \ref{fig:table_fail} shows the failure rates of each method for various $L$ values using the same data used in Figure \ref{fig:MSE_Gi0}. Our networks also overperform both MLE and LCUM on this domain for all depicted scenarios, especially for lower $L$, where the estimation is typically more challenging \cite{frery2022sar}. Counterintuitively, more failures are detected in our methods as $L$ increases. Further experimentation is necessary to better understand this phenomenon, and it is left to future work. We also found that estimation on our networks is generally 3 times faster than the other methods. 
%These numbers, however, considering our non-optimized implementations, have potential for improvement.

\subsection{Results on Real Data}
Figure \ref{fig:real_results} shows the roughness maps estimated by networks trained according to Algorithm \ref{alg:rougness_image} and using kernel sizes $k = \{3, 11, 20, 45\}$ during inference. Note that the $\alpha$ values in all maps follow their expected understanding: higher values are related to urban areas; moderate values suggest forest zones and low values correspond to lake regions \cite{gambini2015parameter}. This demonstrates our methodology's ability to correctly estimate the desired pixel roughness, despite the networks having only been trained on purely synthetic data. Furthermore, we see the blurring effect one expects from applying larger kernels in their convolutions. More interestingly, however, is the observation that the network is still able to predict an expected map using kernel sizes it was not trained one (such as $k = \{20, 45\}$). Finally and more importantly, Figure \ref{fig:real_results} also provides the timings to the computation of each map. Here, we note that we are able to estimate the roughness of each individual pixel of an image as large as $1500 \times 1500$ pixel in less than a tenth of a second for $k = 3$. Processing an image at this rate is usually considered real time \cite{bradski2012real}, meaning that this estimation can be performed as the images are being acquired. As $k$ increases, the map computation becomes slower, but it is still quickly accomplished for low $k$.

\section{Conclusion}
We proposed a neural-network based algorithm for roughness estimation in $G_I^0$ modeled SAR data, a task that is increasingly crucial in SAR image understanding. Practically, it consists of using log-moments from the samples as the input of a network that, when trained, outputs the underlying roughness value. This same network can be easily adapted to process SAR images, from moment computation to parameter estimation, leading to quick estimation of pixel-level roughness. We empirically demonstrate that such networks trained on purely synthetic data are able to outperform traditional estimation methods and return reliable roughness maps from high resolution SAR images in real time. Overall, this result shows that one can use cheap synthetic data to train performant networks for SAR imaging tasks, where labeled data acquisition is expensive. SAR data is in fact ideal for this approach, since there are good statistical models for them that allow the generation of synthetic training data, and they are structured such that important operations on them can be implemented on GPUs. Future work will consist of applying similar techniques to other data domains, where these aspects can also be found.

% \begin{table}[t]
% \caption{Average estimation time (s) on synthetic data.}\label{table_speed}
% \centering
% \begin{tabular}{cccccc}  
% \toprule[1pt]
% \multirow{2}{*}{Model}&\multirow{2}{*}{$n$} & \multicolumn{2}{c}{Proposed Method} & \multirow{2}{*}{\shortstack{Traditional \\ LCUM}}\\
% & & Non-corrected & Corrected &  \\\midrule
% \multirow{ 3}{*}{$G_I^0$} 
% & 9     & 0.56e-04 & 1.13e-04 & 2.28e-03\\
% & 121   & 0.87e-04 & 1.14e-04 & 2.13e-03\\
% & 1000  & 2.33e-04 & 3.43e-04 & 2.17e-03\\\midrule
% \multirow{ 3}{*}{$G_I^0$} 
% & 9     & 0.59e-04 & 1.05e-04 & 2.71e-03\\
% & 121   & 0.91e-04 & 1.37e-04 & 2.16e-03\\
% & 1000  & 2.55e-04 & 3.41e-04 & 2.18e-03\\\bottomrule[1pt]
% \end{tabular}
% \end{table}

\bibliography{mybib}
\bibliographystyle{IEEEtran}

\end{document}